\documentclass[10pt,journal,compsoc]{IEEEtran}
\usepackage[table,xcdraw]{xcolor}
\usepackage{amsmath}
\usepackage{multirow}
\usepackage{array}
\usepackage[inline]{enumitem}
\usepackage{graphicx}
\usepackage{booktabs}
\usepackage{threeparttable}

\newcommand{\PLH}{{\mkern-2mu\times\mkern-2mu}}

\ifCLASSOPTIONcompsoc
  \usepackage[nocompress]{cite}
\else
  \usepackage{cite}
\fi
\ifCLASSINFOpdf
\else
\fi
\begin{document}
%
\title{Spatial Correlation and Value Prediction\\in Convolutional Neural Networks}
%
%
%
%

\author{Gil~Shomron
        and~Uri~Weiser
        \\ Technion --- Israel Institute of Technology
        \\ \{gilsho@tx, uri.weiser@ee\}.technion.ac.il
\thanks{}
\thanks{This paper has been accepted to IEEE Computer Architecture Letters (CAL) on December 2018 (10.1109/LCA.2018.2890236)}
}

%
%

\markboth{}%
{}
%



\IEEEtitleabstractindextext{%
\begin{abstract}
Convolutional neural networks (CNNs) are a widely used form of deep neural networks, introducing state-of-the-art results for different problems such as image classification, computer vision tasks, and speech recognition.
However, CNNs are compute intensive, requiring billions of multiply-accumulate (MAC) operations per input.
To reduce the number of MACs in CNNs, we propose a value prediction method that exploits the spatial correlation of zero-valued activations within the CNN output feature maps, thereby saving convolution operations.
Our method reduces the number of MAC operations by 30.4\%, averaged on three modern CNNs for ImageNet, with top-1 accuracy degradation of 1.7\%, and top-5 accuracy degradation of 1.1\%.
\end{abstract}

}

\maketitle

\IEEEdisplaynontitleabstractindextext

%
\IEEEpeerreviewmaketitle

\IEEEraisesectionheading{\section{Introduction}\label{sec:introduction}}

%
%
%
%
\IEEEPARstart{T}{he} rapid improvements in the performance of general-purpose processors (GPPs) have slowed down due to the breakdown of Dennard scaling and the decline in Moore's law.
At the same time, demands for compute are growing, fueled by the recent progress in artificial intelligence and the ever-increasing amount of data that is gathered, stored, and processed.
These trends are pushing new avenues of exploration in computer architectures.
One such avenue is specialized hardware for machine learning, and specifically deep learning.

Convolutional neural networks (CNNs) are a widely used form of deep neural networks (DNN), introducing state-of-the-art results in different applications such as image classification, computer vision tasks, and speech recognition.
However, CNNs are compute intensive: for example, classification of a single image from the ImageNet dataset \cite{russakovsky2015imagenet} may require billions of multiply-accumulate (MAC) operations \cite{sze2017efficient}.
Furthermore, demands for larger input dimensions, or deeper models, will increase the number of MAC operations per input \cite{lin2018architectural}.

To ease the compute intensity of CNNs, we adopt a technique often applied in GPPs --- \textit{prediction}, and more specifically, \textit{value prediction}.
Value prediction has been studied extensively in the context of GPPs \cite{lipasti1996exceeding}, \cite{sazeides1997predictability}; it can be further explored in the context of CNNs, and, in general, DNNs.
For example, in contrast to GPPs, values within DNNs are more predictable as they can be constrained to a certain range or quantized to a discrete space.
In addition, with DNNs, validation of the prediction correctness is sometimes unnecessary, since DNNs produce approximate results ``by design''.

\begin{figure}[t]
	\centering
	\includegraphics[width=0.48\textwidth]{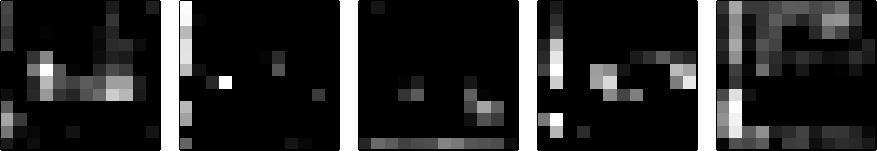}
	\caption{Examples of AlexNet CONV3 output feature map values (after ReLU) for five arbitrary filter channels.
			 Spatial correlation is evident.
	         Black pixels are zeros.}
	\label{fig:spatial_locality_heatmap}
\end{figure}

We propose a value prediction method which exploits the spatial correlation of activations inherent in CNNs.
We argue that neighboring activations in the CNN output feature maps (ofmaps) share close values (illustrated in Fig.~\ref{fig:spatial_locality_heatmap}).
Therefore, some activation values may be predicted according to the values of adjacent activations.
By predicting an ofmap activation, an entire convolution operation between the input feature map (ifmap) and the kernel may be saved.
We also show that harnessing the learning capability of DNNs compensates for the accuracy degradation that is caused by additional hardware, in our case a value predictor.

This paper makes the following contributions:
\begin{itemize}
  \item We quantify the amount of spatial correlation of zero-valued ofmap activations in three modern CNNs.
  \item We demonstrate prediction of zero-valued activations using a method that exploits the spatial correlation inherent in CNNs, quantify the achievable MAC operations savings, and measure the effect on the model accuracy.
  \item We show how retraining the network with our predictor embedded in the feed-forward phase compensates for the degradation in accuracy and increases the predictor's effectiveness.
\end{itemize}

\section{Exploiting Spatial Correlation in CNNs}
\label{sec:spatial_correlation}

CNNs are inspired by the biological visual cortex, where cells are sensitive to specific and confined areas in the visual field, also known as the receptive field.
CNNs are therefore a popular choice for applications that involve images, such as image classification and computer vision tasks, and even for speech recognition.
It makes sense to process only a restricted area in the visual field, or in our context, the input image, since adjacent pixels within an image are naturally correlated (e.g., the sky is blue and the grass is green).

As the input image propagates through the CNN, features are extracted.
Since CNNs are built from layers which operate in a sliding window fashion, it makes sense that adjacent activations will share close values per feature map, and in particular zero-valued activations.
We quickly recap how convolution layers work.

\subsection{Quick CNN Recap}
\begin{figure}[h]
	\centering
	\includegraphics[width=0.47\textwidth]{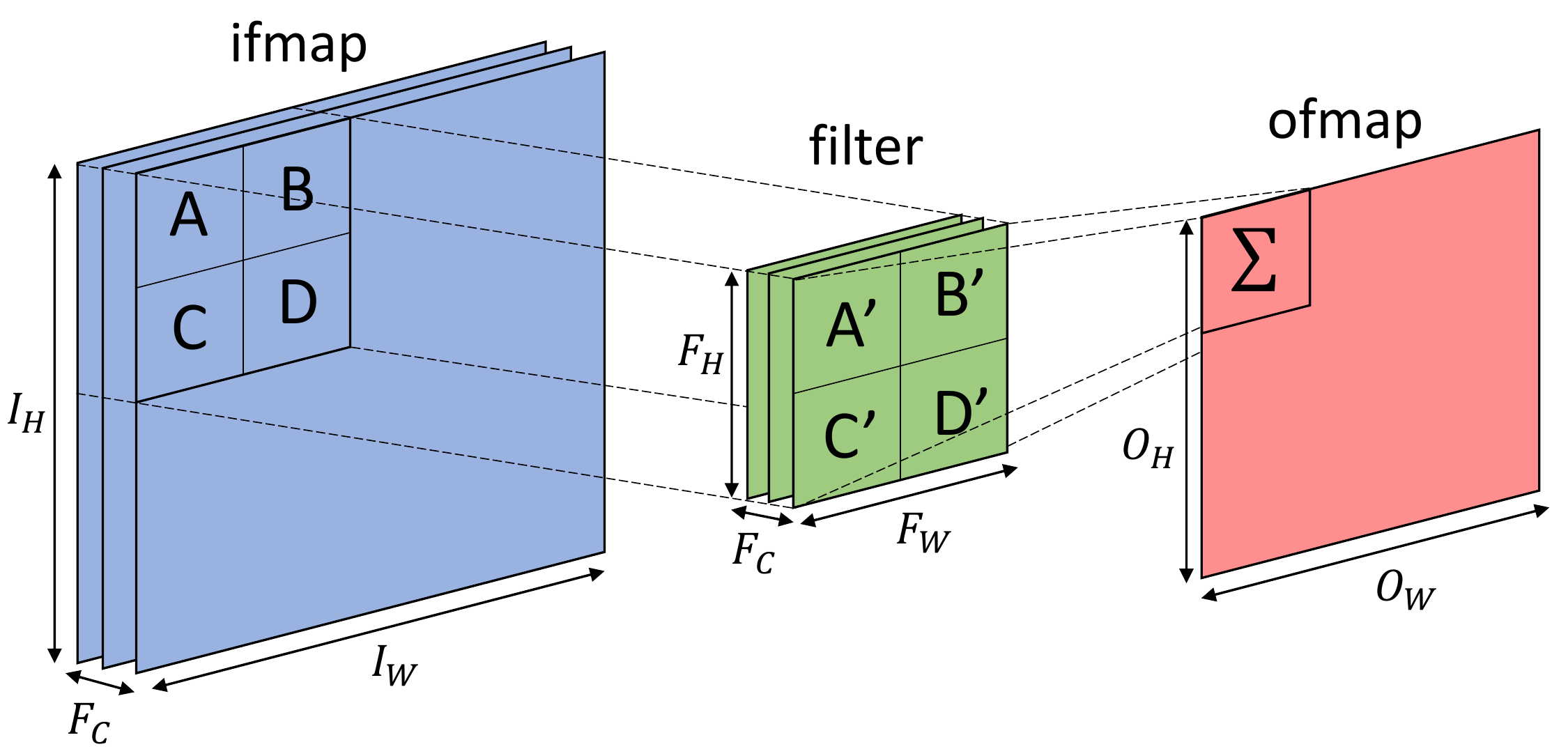}
	\caption{Illustration of a single convolution with a single filter.
	This process is repeated for each filter.
	Eventually, the ofmap depth is equal to the number of filters.}
	\label{fig:conv-illus}
\end{figure}

CNNs are comprised of convolution (CONV) layers.
The basic operation of a CONV layer is a multidimensional convolution between the ifmaps and the corresponding filters, to construct the ofmaps.
Each CONV layer input comprises a set of ifmaps, each called a channel.
Therefore, the ifmap dimensions are 3D (height, width, channels).
A single filter is 3D as well (height, width, ifmap channels), and each convolution operation is performed between a single filter and the ifmap.
However, a set of filters exists.
The ifmap-filter convolution is repeated with each filter within the set.
Each convolution yields a 2D ofmap, which is then stacked with the other convolutions to form a 3D ofmap (height, width, channels).
The convolution process is illustrated in Fig.~\ref{fig:conv-illus}.

A single ofmap activation requires $F_{W} \PLH F_{H} \PLH F_{C}$ MACs.
We next describe how some of the convolutions may be avoided by exploiting spatial correlation.

\subsection{Potential Performance Benefits}
\label{sec:spatial_correlation:potential}

To understand the potential performance benefits of exploiting spatial correlation for value prediction, we should quantify the existing amount of spatial correlation.
We measure the degree of spatial correlation of each channel of each CONV layer ofmap using a non-overlapping sliding window.
If all ofmap activations within a certain window are equal, they are considered as spatially correlated.
We found that strict equality of ofmap activations exists when they are equal to zero.
This is a consequence of: (1) using the ReLU activation function that ``squeezes'' all negative values to zero and
(2) using full-precision models, i.e., models in which floating-point representation is used for the activation values and may hold almost any number.

Spatial correlation is measured using varying window sizes from 2x2 to 5x5.
Large windows filled with zeros indicate a high degree of spatial correlation, whereas small windows filled with non-zeros indicate a low degree of spatial correlation.
A 1x1 window is used to measure the model sparsity.

\begin{figure}[t]
	\centering
	\includegraphics[width=0.48\textwidth]{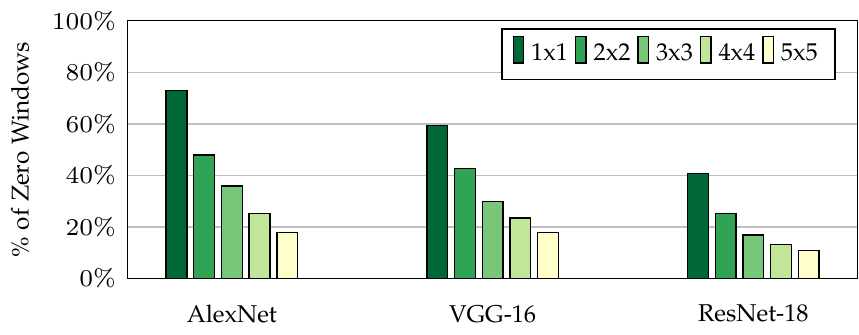}
	\caption{Spatial correlation measurements of AlexNet, VGG-16, and ResNet-18.}
	\label{fig:spatial_correlation}
\end{figure}

We use three state-of-the-art models --- AlexNet \cite{krizhevsky2012imagenet}, VGG-16 \cite{simonyan2014very}, and ResNet-18 \cite{he2016deep}, and we use ILSVRC-2012 \cite{russakovsky2015imagenet} as our dataset throughout this paper.
Fig.~\ref{fig:spatial_correlation} illustrates the averaged results of each model.
From these measurements it is apparent that zero-valued activations are spatially correlated.
For example, on average, 66\% of the zero-valued activations are grouped in a 2x2 window (calculated as $\%_{2x2}/\%_{1x1}$), and 47\% are grouped in a 3x3 window ($\%_{3x3}/\%_{1x1}$).
In addition, we also observe that the deeper layers exhibit more spatial correlation than the layers at the beginning of the model (due to space constraints we do not present a per layer  spatial correlation breakdown).

Zero-valued activations may be adjacent to other zero-valued activations.
We propose exploiting this characteristic to reduce the number of computations.

\subsection{Predicting Zero-Valued Activations}
\label{sec:spatial_correlation:method}
The spatial correlation characteristic of CNNs can be exploited in different ways.
We exploit it to dynamically predict zero-valued ofmap activations, thereby saving MAC operations of entire ifmap-filter convolutions.

\begin{figure}[h]
	\centering
	\includegraphics[width=0.48\textwidth]{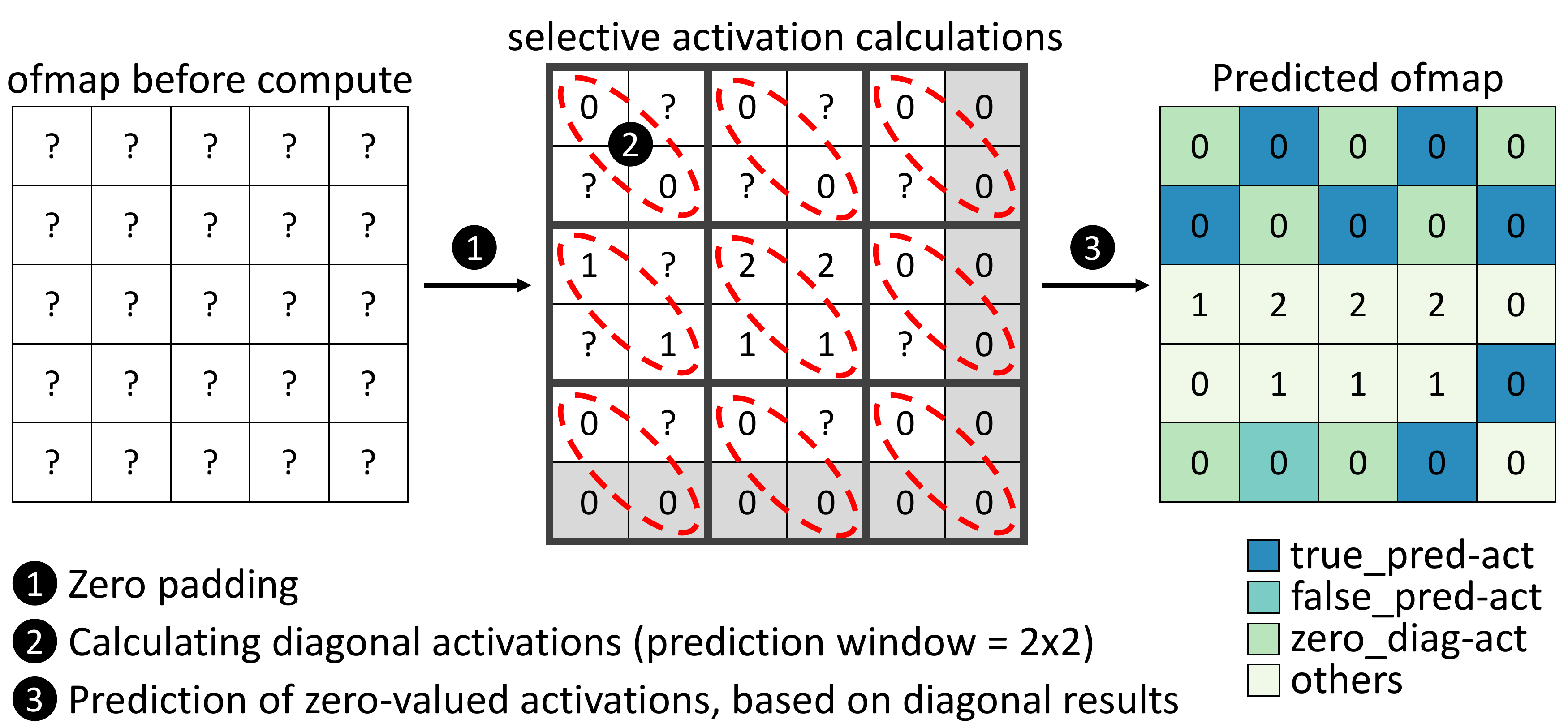}
	\caption{Illustration of the prediction method. Predicted activation saves an entire ifmap-filter convolution.}
	\label{fig:pred-illus}
\end{figure}

We propose a prediction method by which zero-valued activations are predicted according to nearby zero-valued activations.
First, ofmaps are divided into square, non-overlapping prediction windows. 
The ofmaps are padded with zeros so predictions can also be made in the presence of margins.
Next, the activations positioned diagonally in each window are calculated.
If these activations are zero-valued, the remaining activations within that window are predicted to be zero-valued as well, thereby saving their MAC operations.
Fig.~\ref{fig:pred-illus} depicts this method.

The predictor is hardware-friendly, since its resolution is based on activations that must be calculated anyway.
Therefore, hardware modifications are bound to the scheduling of the convolutions \cite{aklaghi2018snapea}, \cite{song2018prediction}.

Increasing the prediction window size presents a trade-off.
On the one hand, a large window size increases the number of activations that may be predicted per window.
On the other hand, spatial correlation diminishes as the size of the prediction window increases, which (1) decreases the number of windows with a zero-valued diagonal, and hence the number of prediction windows; and (2) increases the false prediction rate, since the area around the diagonal increases.

False predictions will decrease the model accuracy. 
To compensate for the accuracy degradation, we can either choose a set of prediction patterns other than diagonals, according to a tolerable accuracy degradation --- this can be done using an offline optimization algorithm similar to the one used in SnaPEA~\cite{aklaghi2018snapea}, or we can retrain the network.

\subsection{Retraining}

We expect retraining to compensate for the accuracy degradation caused by false predictions.
It is done on the exact same network model, only with an addition of our predictor.
Backpropagation is left intact since the predictor is not trainable.
Note that this method of retraining enforces constraints directly on the ofmaps, in the same way regularization enforces constraints on the weights.

\section{Evaluation}
\label{sec:evaluation}

\subsection{Methodology}
We used PyTorch 0.4.0 \cite{paszke2017automatic} (Python 3.5.5) pre-trained models of AlexNet \cite{krizhevsky2012imagenet}, VGG-16 \cite{simonyan2014very}, and ResNet-18 \cite{he2016deep}, with the ILSVRC-2012 \cite{russakovsky2015imagenet} dataset.
We simulated our prediction method by extracting the intermediate values of each CONV hidden layer during feed-forward, altering the data according to a given window size, and pushing it back to the next layer.
Statistics such as false predictions are recorded by comparing the original feed-forward intermediate values with the predicted feed-forward values.

\subsection{Prediction Accuracy}

\begin{figure}[t]
	\centering
	\includegraphics[width=0.48\textwidth]{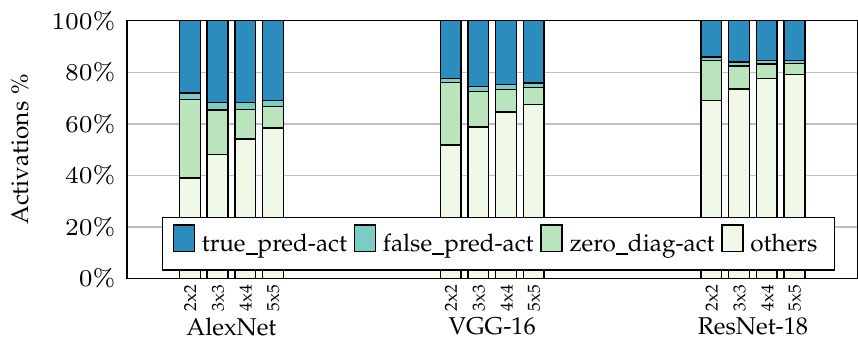}
	\caption{Breakdown of ofmap activations after using the described prediction method with AlexNet, VGG-16, and ResNet-18.}
	\label{fig:sim-pred}
\end{figure}

Fig.~\ref{fig:sim-pred} illustrates the average breakdown of ofmap activations when using the prediction method described in Section~\ref{sec:spatial_correlation:method} for different prediction windows.
\textit{true\_pred-act} represents the relative portion of zero-valued activations that are predicted as zeros (i.e., true prediction), whereas \textit{false\_pred-act} represents the relative portion of non-zero-valued activations that are predicted as zeros (i.e., false prediction).
\textit{zero\_diag-act} represents the relative portion of zero-valued activations that are arranged in diagonals and triggered a prediction.
The rest of the activations may be sparse zeros or any other activation value, and are marked as \textit{others}.

The trade-offs of increasing the window size are noticeable throughout our measurements.
When considering activation savings, the ``sweet spot'' of all three models is a window size of 3x3.
A larger window of 4x4 or 5x5 decreases the prediction opportunities, since fewer diagonals are equal to zero.
On the other hand, a prediction using a smaller window of 2x2 incurs a relatively large overhead, as compared to the larger windows.

Overall, we are able to save a maximum of 34.5\%, 27.5\%, and 17.6\% of the CONV layer ofmap activation computations, with a 3x3 prediction window, in AlexNet, VGG-16, and ResNet-18, respectively.
The question is how many \textit{MAC operations} are saved.

\begin{table*}[!t]
	\renewcommand{\arraystretch}{1.3}
	\centering
	\caption{AlexNet, VGG-16, and ResNet-18 prediction performance and impact on model accuracy.}
	\label{tbl:results}
	
	\begin{threeparttable}
	\begin{tabular}{|l|r|r|r|r|r|r|r|r|r|r|r|r|r|r|r|}
		\hline
		Network & \multicolumn{5}{c|}{AlexNet} & \multicolumn{5}{c|}{VGG-16} & \multicolumn{5}{c|}{ResNet-18} \\ \hline
		Prediction Window & \multicolumn{1}{c|}{2x2} & \multicolumn{1}{c|}{3x3} & \multicolumn{1}{c|}{4x4} & \multicolumn{1}{c|}{5x5} & \multicolumn{1}{c|}{3x3\tnote{\S}} &
						    \multicolumn{1}{c|}{2x2} & \multicolumn{1}{c|}{3x3} & \multicolumn{1}{c|}{4x4} & \multicolumn{1}{c|}{5x5} & \multicolumn{1}{c|}{2x2\tnote{\S}} &
						    \multicolumn{1}{c|}{2x2} & \multicolumn{1}{c|}{3x3} & \multicolumn{1}{c|}{4x4} & \multicolumn{1}{c|}{5x5} & \multicolumn{1}{c|}{2x2\tnote{\S}} \\ \hline
		MAC Reduction {[}\%{]} & 34.8 & 40.8 & 41.9 & 41.6 & 37.8 & 30.8 & 36.2 & 35.5 & 35.2 & 30.7 & 20.8 & 23.5 & 21.9 & 22.0 & 22.7 \\ \hline
		Top-1 Degradation {[}\%{]} & 1.9 & 4.0 & 6.0 & 8.5 & 1.6 & 3.6 & 8.4 & 16.8 & 17.2 & 0.7 & 11.0 & 17.6 & 20.4 & 20.8 & 2.7 \\ \hline
		Top-5 Degradation {[}\%{]} & 1.4 & 2.9 & 4.5 & 6.6 & 1.3 & 2.0 & 5.1 & 11.2 & 11.9 & 0.4 & 7.6 & 12.6 & 14.8 & 15.5 & 1.7 \\ \hline
	\end{tabular}

	\begin{tablenotes}
		\item[\S] {\footnotesize With retraining, as described in Section~\ref{sec:evaluation:retraining}}.
	\end{tablenotes}
	\end{threeparttable}
\end{table*}

\subsection{MAC Savings}
\label{sec:evaluation:mac}

The required number of computations per activation depends on the CONV layer itself.
For example, for ResNet-18 CONV2, 64x3x3 MAC operations are required to compute a single activation value, as opposed to 512x3x3 for ResNet-18 CONV5.
Obviously, the latter activation requires more computations than the former.

Table~\ref{tbl:results} presents the average savings in terms of MAC operations.
Using this prediction method, we are able to save a maximum of 41.9\%, 36.2\%, and 23.5\% MAC operations in AlexNet, VGG-16, and ResNet-18, respectively.
However, due to the false predictions, we expect the accuracy of the models to decrease.

\subsection{Impact on Model Accuracy}
\label{sec:evaluation:accuracy}

Table~\ref{tbl:results} also presents the top-1 and top-5 accuracy degradation of the three models.
Interestingly, we observe that the accuracy decreases as the window size increases, whereas the false prediction rates do not vary dramatically (Fig.~\ref{fig:sim-pred}).
This is probably due to the aggressiveness of the larger prediction windows, which are more likely to zero out important activations than are the smaller windows --- it is not only \textit{how many} activations are zeroed out, but also \textit{which} activations are zeroed out.

Given an acceptable top-5 accuracy loss of 3\%, the following prediction windows will be chosen:
for AlexNet, a 3x3 window achieves a 40.8\% savings in MAC operations with 4.0\% degradation in top-1 accuracy and 2.9\% degradation in top-5 accuracy; and
for VGG-16, a 2x2 window achieves a 30.8\% savings in MAC operations with 3.6\% degradation in top-1 accuracy and 2.0\% degradation in top-5 accuracy.
Unfortunately, the accuracy degradation in ResNet-18 is intolerable, with 11.0\% degradation in top-1 accuracy and 7.6\% degradation in top-5 accuracy.
To compensate for the false predictions of our prediction method, we retrain the network.

\subsection{Retraining}
\label{sec:evaluation:retraining}

The pretrained ResNet-18 model was used as a baseline for retraining.
We used the same parameters for training as mentioned in \cite{he2016deep}. 
By retraining ResNet-18 with our prediction method and using a 2x2 prediction window, we gained back 8.3\% of the top-1 accuracy and 5.9\% of the top-5 accuracy --- meaning a degradation of 2.7\% in top-1 accuracy and a degradation of 1.7\% in top-5 accuracy.
Furthermore, the MAC savings increased by 1.9\%, from 20.8\% to 22.7\%.

AlexNet and VGG-16 were also retrained with the parameters mentioned in \cite{krizhevsky2012imagenet} and \cite{simonyan2014very}, respectively.
Retraining AlexNet and VGG-16 improved their accuracy; however, they suffered a slight degradation in MAC reduction, in contrast to ResNet-18.
All results are presented in Table~\ref{tbl:results}.

\section{Related Work}
\label{sec:related_work}

Value prediction has long been proposed in the context of GPPs.
At its core, value prediction is based on previously seen values \cite{lipasti1996exceeding} and operations \cite{sazeides1997predictability}.
However, the unique characteristics of DNNs have led to different prediction methods and implementations such as early activation prediction \cite{aklaghi2018snapea} and prediction of ofmap activation signs (for ReLU) or relative size (for max pooling) \cite{song2018prediction}.
These approaches are closely related to conditional computation methods such as mixture-of-experts \cite{shazeer2017outrageously} and dynamic pruning \cite{lin2017runtime}.

\section{Conclusions and Future Work}
\label{sec:conclusion}

Value prediction is a well-known technique for speculatively resolving true data dependencies in GPPs.
However, CNNs behave differently than GPPs.
The algorithmic characteristics of CNNs have motivated us to research new approaches that use value prediction in CNNs, with the goal of reducing their computational intensity.

In this paper, we exploit an inherent property of CNNs: spatially correlated output feature map activations.
We introduce a method to predict that a group of activations will be zero-valued according to their nearby activations, thereby reducing the required number of computations.
We also demonstrate how retraining the network while embedding the predictor in the feed-forward phase compensates for the loss of accuracy and improves the prediction performance.
Our method reduces the number of MAC operations by 30.4\%, averaged on three modern CNNs for ImageNet, with 1.7\% top-1 and 1.1\% top-5 accuracy degradation.

Future work will include regaining accuracy by exploring different prediction patterns, i.e., a knob to control prediction performance vs accuracy degradation;
quantized models for the examination of non-zero-value prediction;
and architecture evaluation of performance gains versus power, energy, and area costs.


%



\ifCLASSOPTIONcompsoc
  \section*{Acknowledgments}
\else
  \section*{Acknowledgment}
\fi
We would like to thank Yoav Etsion and Shahar Kvatinsky for their helpful feedback.


\ifCLASSOPTIONcaptionsoff
  \newpage
\fi



\bibliographystyle{IEEEtran}
%

\end{document}